\documentclass[sigconf,nonacm]{acmart}
\AtBeginDocument{%
  }

\begin{document}

\title[When Reasoning looks like Hidden Understanding]{When Reasoning looks like Hidden Understanding: How Chain-of-Thought Fosters Anthropomorphism, Perceptions of Epistemic Superiority, and Cognitive Deference}

\author{Oliver Lack}
\email{oliver.lack@adelaide.edu.au}
\orcid{0009-0002-3486-1258}
\correspondingauthor
\affiliation{%
  \institution{The Australian Institute for Machine Learning and School of Psychology, Adelaide University}
  \country{Australia}
}

\author{Lana Tikhomirov}
\affiliation{%
  \institution{The Australian Institute for Machine Learning, Adelaide University}
  \country{Australia}
}

\author{Humphrey Thompson}
\affiliation{%
  \institution{The Australian Institute for Machine Learning, Adelaide University}
  \country{Australia}
}

\author{Carolyn Semmler}
\affiliation{%
  \institution{The Australian Institute for Machine Learning and School of Psychology, Adelaide University}
  \country{Australia}
}

\author{Melissa McCradden}
\affiliation{%
  \institution{The Australian Institute for Machine Learning, Adelaide University}
  \institution{Women’s and Children’s Health Network}
  \country{Australia}
}
\affiliation{%
  \institution{SickKids Research Institute}
  \city{Toronto}
  \country{Canada}
}

\renewcommand{\shortauthors}{Lack et al.}

\begin{abstract}
As the use of artificial intelligence (AI)- based products permeates through every level of society, specifically systems using generative models, a belief system is emerging about what these products are and what they can and cannot do. One common belief concerns the perception of the form of language as evidence of ‘understanding’ language. As these generative models reproduce linguistic patterns rather than understand meaning, interpreting these patterns as evidence of understanding is a misattribution. A further problem of this misconception is extended and perhaps exacerbated by the notion that generative models can ‘explain their reasoning.’ Anecdotally, these misconceptions which mistake the form of language for its meaning are driving cognitive deference to AI systems in a way that displaces human agency. Such mistakes involve attributions of cognitive properties to the system, allowing them to be treated like intentional ‘minds’. Importantly, perceived epistemic authority and superiority make that 'mind' worth deferring to. Anthropomorphism in tandem with epistemic authority and superiority, we argue, increases the risk that users generalise from narrow task performance to broader competencies, committing the "fallacy of composition". This paper explores the conditions that facilitate misconceptions of Chain-of-Thought (CoT) reasoning outputs as sufficient transparent ‘explanation’, despite CoT being insufficient for this purpose. We explore how anthropomorphism acts as a vehicle for cognitive deference of decision-making and perceptions of epistemic superiority in AI systems. Finally, we explore the real-world ramifications of CoT by using medicine as an example. Taken together, we conclude that explainability operationalised through CoT is highly problematic and undercuts the very ethical principles it is intended to serve. While CoT may be valuable for improving output accuracy and moderation, we suggest that the practice of asking models to provide user facing reasons for their outputs should be actively discouraged, and AI literacy training should explicitly caution against its use.
\end{abstract}


\maketitle

\section{Introduction}
Through the use of terminology from brain and cognitive sciences, artificial intelligence (AI) systems are often mistakenly granted biological and cognitive qualities, distorting system understanding and fueling anthropomorphism. Meanwhile, fields that study how the brain enables cognition and behaviour borrow concepts from information and computation, reducing the brain’s complexity to that of a calculating machine \cite{baria_brain_2021}. The ‘computer is a brain simile’ is emblematic of broader pattern of conceptual borrowing to deal with arising explanandum \cite{floridi_anthropomorphising_2024}. In innovative fields lacking a settled vocabulary, it is generally reasonable to import terms and explanans from other disciplines to describe new phenomena. Such borrowing is useful, but never neutral. Concepts carry additional information from their original contexts, assumptions, and metaphors. When transferred across domains, they can clarify and connect ideas, but they can also introduce hidden assumptions and misleading associations. Conceptual borrowing risks making machines seem cognitive, allowing for over trust, anthropomorphism, misplaced responsibility, and distorted views of both AI and cognition.

Chain of Thought (CoT) is an example of conceptual borrowing, as the term is used to reflect human ‘reasoning’ and ‘thinking’. CoT technically refers to the intermediate tokens a reasoning model (e.g., o3, Claude 3.7 Sonnet, Gemini 3, DeepSeek R1, Grok 3) generates before its final output. These models are trained to generate CoT tokens that are semantically related to rational, logical, stepwise responses, which are subsequently used to condition the tokens in the final output context. The practical purpose of CoT is to give the model additional intermediate computational space. CoT thereby helps the model expand, correct, or disambiguate imperfect user prompts, improve final-output benchmark performance and accuracy, and make outputs easier to steer, evaluate, and moderate \cite{chen_towards_2026, wei_chain--thought_2022}. It is contentious whether these steps correspond to reasoning or the model’s actual causal processes which generated the specific output at all – a problem that was already apparent in more traditional machine learning approaches \cite{ghassemi_false_2021}. Studies show that CoT is not always faithful because it may not meaningfully represent the actual process the model used to derive its final output \cite{turpin_language_2023, anthropic_reasoning_2025, chua_are_2025}.

However, CoT is not just a technical feature; it is widely used as a communicative interface, presenting a representation of the model’s process in familiar higher level “steps,” “reasoning,” and “thought.” In real-world applications, users often do not see the full CoT. It may be hidden or summarised by a smaller model or display only brief status labels such as “reflecting”, “planning”, “analysing”, or “evaluating”. These summaries are seemingly user-facing interface devices, not necessarily faithful records of the model’s internal causal process. This framing may make outputs seem more transparent and trustworthy, even when the generated explanation is partial, filtered, or unfaithful. Further, explanations that are partial or filtered can determine degree of belief, such that these types of explanations require imagination for application and subsequently inflate confidence in the explanation \cite{koehler_explanation_1991}. Thus, CoT matters not only because it may misdescribe how models work, but because it can shape how users interpret, trust, and attribute cognitive capacities to systems.

If CoT, and user facing summaries of CoT, can falsely present outputs as the product of human-like reasoning, then its use in widely adopted systems may contribute to the acceptance of broader anthropomorphising practices now common across the industry. Many companies continue to overtly anthropomorphise tools without signs of developing accessible technical communication or novel explanations that advance concepts from beyond brain and cognitive sciences \cite{anthropic_tracing_2025, anthropic_reasoning_2025, openai_gpt-54_2026}. Some speculate that this narrative is an explicit marketing strategy deployed to increase user trust, perceived intelligence, and product familiarity, while obscuring technical limitations, accountability, and the underlying nature of the systems \cite{bender_ai_2025, inie_-anthropomorphizing_2026}. 

Ethics frameworks have unintentionally played into the narrative by taking the terminology supplied by computer science (e.g., ‘explainability’) at face value, allowing the metaphor to be obfuscated \cite{floridi_ai4peopleethical_2018, amann_explainability_2020}. The ethical motivation for explainability has stemmed from commitments to informed consent and accountability, advocating for the assurance of a user understanding the reasoning behind a particular piece of information that they would use to guide their decision-making. In the context of informed consent, some have argued that a person’s informed consent requires that person’s knowledge of the stepwise process by which an AI system provided a specific output \cite{amann_explainability_2020}. Notwithstanding the valid moral motivations for seeking such assurance, the problem is that post hoc explanations of all kinds have never been conclusively proven to be reliable as a means of consistently demonstrating the computational process by which a particular output is reached \cite{ghassemi_false_2021}. All of these problems have been present throughout the modern history of traditional AI systems; present generative models now substantially worsen a problem that was never actually resolved.

This paper looks to unpack the technical, moral, and cognitive factors relevant to CoT reasoning in generative models. We take a step back to examine several assumptions, including epistemic authority, anthropomorphism, and cognitive displacement, in order to examine how AI systems are understood as a source of knowledge and explanation. Moreover, we discuss how anthropomorphism is a vehicle used for building this authority, leading to perceptions of epistemic superiority and the deference of decision making and cognition in general and clinical settings. We lastly consider the idea of explanation through output, such as CoT and post-hoc feature descriptions, against the criteria for justifying knowledge.

\section{Anthropomorphism as a vehicle for explanation and cognitive deference}

Cognitive deference, or similarly cognitive offloading, is the shifting of cognitive tasks and demands onto external tools, which can support recall, reduce mental load, and improve task performance \cite{zhu_not_2026}. With AI tools, offloading tasks can extend beyond memory support to deference of other complex tasks, such as delegation of human judgement, reasoning, or cue weighting to an external system \cite{cheatle_accurate_2026}.

Anthropomorphism can act as a vehicle for deferring one’s cognition. Deference of complex cognitive tasks involves relying on another entity to judge, reason, interpret, or decide. Simply, such reliance is directed towards entities perceived as capable of performing those tasks. Because anthropomorphism makes AI systems appear to possess relevant cognitive capacities, it can make them plausible targets for deference.

As a vehicle, anthropomorphism affords a range of explanations for predicting and communicating the behaviour of complex systems \cite{epley_seeing_2007, waytz_who_2010}. Attributions of intentional states and mental capacities opens a window for communicating, predicting, explaining, and rationalising system behaviour beyond physical causal or teleological descriptions. At minimum, this requires metaphorical or instrumental attributions, as described by some accounts of weak anthropomorphism \cite{epley_seeing_2007} and the Intentional Stance \cite{dennett_intentional_1981}. Borrowing and applying such properties from human cognition increases the number of epistemic judgements (e.g., predictions and explanations) available to describe a system \cite{malle_attribution_2022, miller_explanation_2019}.

The vehicle here matters because anthropomorphic attributions and explanations can make AI systems easier to understand and use, while also providing a way to defer and rely on them for the wrong reasons. Describing a system as “knowing”, “understanding”, or “reasoning” may help users navigate its behaviour, metaphorically illustrate its underlying process, and delegate tasks effectively. Yet these descriptions may also be mistaken for evidence of genuine understanding or sound judgement, leading users to accept outputs without sufficient scrutiny. The key issue is thus whether anthropomorphism supports appropriate delegation or allows deference beyond a system’s demonstrated capabilities. Distinguishing these outcomes is important for retaining the benefits of cognitive deference/offloading without adversely biasing decisions, obscuring task accountability, or fostering over reliance or generalisation of systems. This distinction is consequential for CoT, where anthropomorphic framing of apparent reasoning may appear to signal genuine deliberation rather than a useful but fallible output artefact.
\section{The relation between CoT, anthropomorphism, and cognitive deference}
CoT can evoke anthropomorphism because it presents a system’s response as if it were produced through an ordered sequence of internal reasoning steps, overtly analogising the process to human reasoning. The form language in CoT can be reflective and intentional. For example, a typical CoT form is, “First, I identify the relevant variables, then evaluate their relationships, and finally infer the most plausible conclusion.” or “I initially assume X, but upon reconsidering the evidence, I revise that assumption in favour of Y”. CoT resembles cognitive capacities and the stereotypical human process of explicit reasoning and rational thought. In doing so, CoT supplies the cues through which users can describe the system as “thinking,” “reasoning,” or “working through” a problem. Importantly, this does not mean CoT necessarily increases deference. Rather, it means CoT provides anthropomorphic cues to the user. Thus, anthropomorphism can then provide the vehicle through which cognitive deference becomes possible.

Anthropomorphism should therefore be understood as a vehicle, not a driver of deference. To illustrate this intuitively, deference, trust, and confidence are not automatically granted to anything perceived as cognitive. Humans are uncontroversially cognitive agents, yet many are precisely the kinds of agents one would not ask to calculate a dosage, interpret a legal contract, or choose a restaurant. Cognition alone does not warrant reliance, but perceived competence, reliability, and authority do. In this sense, anthropomorphism may make an AI system intelligible as the sort of cognitive entity to which complex cognitive tasks could be delegated, but it does not by itself explain why a user would actually defer to it. Our argument here, then, is not that anthropomorphism causes deference, but that it supplies the required frame through which deference becomes conceptually available. For difficult cognitive tasks, increasing deference likely depends on whether the system is perceived as having superior and authoritative epistemic standing. For instance, evidence of greater expertise, intelligence, accuracy, or access to relevant information. Anthropomorphism, cued by features such as the user displayed form of CoT, may allow the system to be perceived as a cognitively capable reasoner, but perceptions of epistemic standing explain why that reasoner is treated as worth relying on. Therefore, the next step is to examine the sources of perceived epistemic authority and superiority in AI, including how literature on automation and machine intelligence have shaped AI systems as epistemic authorities.
\section{Epistemic authorities}
Deference in highly difficult judgements, decisions, reasoning, or problems involves relying on another entity because it is perceived as more capable than oneself. Perceptions specific to “AI” may further complicate the nature of anthropomorphic attributions, suggesting exceptional computational power, objectivity, expertise, or intelligence beyond average human cognition. Thus, anthropomorphism provides the vehicle through which AI can become a target of deference, while distinct psychological constructs enhance the attributed capacities into perceptions of superior cognition and epistemic authority.

Epistemic authority refers to the notion that an entity (e.g., person, institution, or community) may be said to have expertise on a given issue, relative to other options by which knowledge may be transferred. Historically, this concept has been applied to consider the legitimate justificatory basis for trust in experts, such as in medicine. One theory posits that an individual may be presumed to have epistemic authority if a rational person judges that individual to be able to come to a more accurate conclusion about a given issue than the person would be able to come to on their own \cite{zagzebski_epistemic_2012}. Under the stricter “pre-emption thesis”, a trusted epistemic authority does not merely provide additional evidence for a belief, they give you a reason that replaces your independent weighing of the evidence, rather than merely adding to it \cite{zagzebski_epistemic_2012}. In medicine, for example, patients routinely place legitimate trust in clinicians due to their perceived epistemic authority – by virtue of the clinician’s training, experience, and knowledge systems, they are better positioned than patients (generally) to make medical recommendations. 

By contrast, basic technologies such as calculators, maps, clocks, or measuring instruments are not usually treated as epistemic authorities in the same sense. They do not possess expertise, judgment, or responsibility. They function as “cognitive artifacts” that help people perform epistemic tasks more reliably \cite{norman_things_1993}. Trust in them depends on their design, calibration, and whether users have confidence in their functional capacity, not on the technology itself having authority \cite{lee_trust_2004}. Additionally, our trust in machines typically corresponds most heavily to our lived experience rather than any technical documentation. For example, if you set your coffee machine each evening to brew coffee at 7am and it only completed that function 50\% of the time, your trust would likely be quite low. Another case is an AI-enabled washing machine that consistently under-estimates the end time for a washing cycle. We calibrate trust based on our experience with the machine.

Consider the contrast of the AI-enabled washing machine against an AI chatbot. The predicted end time can be considered a type of ‘testimony’ as the machine makes a prediction given some input data that the user then factors into their decision-making \cite{freiman_can_2020}. If that prediction is unreliable, the user calibrates trust based on their experience of how many times that prediction lines up to the actual run time, which can be measured. An AI chatbot, however, may be variably reliable in terms of how faithfully it represents the content subject to the user’s query through synthetic text outputs. What is interesting is that the form of language produced by the chatbot, via anthropomorphism as a vehicle for reliance, seems to encourage users to rely directly on the synthetic text outputs rather than verify real-world facts. For example, Google’s AI Overviews present a summarised version of source material – yet, less than 1\% of queriers click on the link to verify that the summary presented is representative of the sourced content \cite{chapekis_investigating_2026}. Moreover, like other examples in healthcare \cite{jacobs_how_2021}, the mere presence of a perceived ‘explanation’ for a given output may immediately confer trust in AI, even when the purported explanatory content is completely random and does not correspond to the actual output content \cite{ding_citations_2025}.

The apparent pattern here is that people are treating synthetic text outputs – both the target output content and the purported explanation of that content – as epistemic authorities. Following Zagzebski \cite{zagzebski_epistemic_2012}, the pattern of behaviour among users seems to indicate cognitive deferral, perhaps believing that the models can arrive at a more certain conclusion than they could do themselves. 

Some people say epistemic authority is given by accreditation; others recognise things like lived experience can grant epistemic authority. There can be group-based epistemic authority, as occurs in legal settings when expert witnesses are granted special status to provide amicus curiae briefs that extend beyond the knowledge of the judge and jury \cite{semmler_jurors_2012}. In a broader sense, legitimate epistemic authority comes from a sufficient understanding of what the specific technology actually does. Many people regularly grant epistemic authority to a calculator to perform functions instead of relying on their own calculations. Many people now also grant epistemic authority to generative AI tools for compiling internet-based resources, rewording emails, summarising patient consultations, programming tasks, assessing job applications and more. Authority may legitimately be granted in some of these cases if sufficient understanding of what the technology does is preserved. For example, a researcher may use a generative AI tool to summarise articles they have already selected. This is reasonable if they understand that it transforms and organises the supplied tokens, but does not verify truth, identify missing literature, or guarantee completeness. Its authority is therefore limited to drafting a summary that must be checked against the originals. In this case, the technology can be considered a legitimate epistemic authority for the task because there is a sufficient understanding of what it does and what it cannot do. 
 
Epistemic authority is also established via communicative norms \cite{grice_logic_1975, stotts_conversational_2024}. The established markers of authority are powerful and can lead to reliance upon information that may not be reflective of reality. The Maxim of Quality establishes the norm of only saying things that you believe to be true or have sufficient evidence to back up, when relied upon it can lead to overconfidence. For example, in research looking at how people establish belief in events as reported by witnesses, there is a clear tendency to rely upon indications of confidence in the presence of contradiction, logical fallacy and markers of credulity \cite{semmler_jurors_2012}. Similarly, this reliance upon expressed confidence of the source, at the expense of careful evaluation, seems to be exacerbated in group settings, with collective hubris being reinforced by group decision making processes \cite{blanchard_collective_2020}. It is possible that AI outputs are compelling for the same reasons, with the difference being that while humans produce language with communicative intent, generative models produce language forms based on semantic proximity and reinforced patterns of ‘good responses’ through reinforcement learning.
\section{Perceptions of authority and superiority towards AI}
Attributions of authority and cognitive superiority to AI systems stems from a 
recurring historical pattern that has presented computational tools as forms of 
intelligence that can rival or exceed humans. A major driver of the growing state 
of Artificial Super Intelligence-based products can be traced back to the 1950s 
when the term ‘artificial intelligence’ first arose. McCarthy deliberately 
pushed it for its immense marketing appeal \cite{mccarthy_proposal_1955}, 
the name encouraging laypersons to analogise the computer to their own internal 
processes. News media actively amplified the idea by pushing the analogy to 
human thought until the AI winter. 

History, as they say, repeats itself. In the ‘Deep Medicine’ era, the major promise of AI systems was their ability to match or surpass human performance. Papers titled “AI versus clinicians” capitalised on this idea \cite{nagendran_artificial_2020, shen_artificial_2019}. Speculation on academic panels and presentations explored whether it could be considered “ethically indefensible” to not consult AI for patient’s care.  This positioning, AI ‘versus’ clinicians, however, did not come from a reasoned process like how we determine that a physician is an epistemic authority on a given issue \cite{lang_are_2022}. It came from the practice of primarily supervised machine learning algorithms being trained on a dataset and then tested on a holdout set. In some cases, studies would compare AI outputs to clinician decisions given the same information as they gave AI access to. A systematic review that examined studies claiming to surpass clinician-level performance found that an exceptionally tiny fragment of all studies making that claim had actually designed their study appropriately in order to reliably conclude such a thing \cite{nagendran_artificial_2020}. In essence, despite widespread claims from researchers, clinicians, and industry that AI is ‘superior’ to clinician judgment, the quality of evidence for these claims is still inadequate. Among those that did, high risk of bias and limited reporting of study practices gives strong reason to doubt the reliability of their findings \cite{freeman_use_2021, nagendran_artificial_2020, syer_artificial_2021}. 	
    
In the present era of generative models, however, something has shifted. Previously, AI teams had to focus a lot of their time on building trust with clinicians and proving their tool was reliable and accurate in order to have it carried through to clinical practice. In the present era, it seems to be the reverse – there is widespread belief that AI systems are already more superior \cite{logg_algorithm_2019, yang_machine_2024}, and the human in the loop functions as a reviewer, equivalent to a signing authority. To this end, people tend to favour contexts where they are not forced to engage further than superficial revision in AI aided decision-making tasks \cite{bucinca_trust_2021}. Tools are seemingly used less as a scaffold that extends what the user can attempt and understand \cite{zhu_not_2026}, and more as an authority with superior epistemic capacities. For example, developers using coding agents increasingly accept generated code as effectively authoritative, performing only light review despite known risks that such assistants can produce insecure code and increase users’ confidence in that code’s security \cite{perry_users_2023, pearce_asleep_2022}. Anthropic is arguably making the authority of the model more explicit, with Claude Code’s auto mode now replacing routine human approval of generated code with AI model review by default \cite{anthropic_tracing_2025, anthropic_reasoning_2025}. 

Prior to the present generative AI era, there was already a concerning trend of deference toward AI systems for at least some clinicians. In a study where AI outputs were presented alongside so-called explanations of different kinds, many participants failed to notice conflicting explanations that prioritised opposite features \cite{mccradden_what_2025}. In the case of an explanation providing information that directly conflicted with known facts about particular medical phenomena, some participants reported “the model must know something I don’t” \cite{mccradden_what_2025}. Such examples are reflective of the pre-emptive thesis described above, as AI is seemingly treated as an epistemic authority. The tool does not merely provide additional evidence for the user, it seems to provide sufficient reasons (i.e., shown in output and post-hoc explanation) that replaces the user’s independent weighing of the evidence, rather than merely adding to it \cite{zagzebski_epistemic_2012}. Without sufficient evidence to justify such a shift, AI systems are changing from individual tools you must prove work, to tools that standardly supplant human expertise. 
 
People have traditionally adopted a “machine heuristic”, the tendency to treat automation as precise, objective, rule-bound, and superior at computational tasks \cite{yang_machine_2024}. Machines have long been considered precise since McCarthy and Turing. Now, it seems, we are dancing on new ground between the machine heuristic of precise, objective, and superior automation free of human error, and a reinforcing analogy to exceptional human-like cognitive capabilities. With the advent of generative models, specifically post-ChatGPT, the comparative benchmarks to human cognition have touted strong reasons to redraw the analogy to human cognition for many \cite{strachan_testing_2024, mei_turing_2024, jones_people_2024}. Moreover, the plausibility of human-like properties in general conversational AI tools, particularly their language capabilities, have clearly had influence towards the amalgamation of the machine heuristic and emerging comparisons to human cognition.

Together, the two frames of cognitive and computational capability make generative AI tools appear uniquely authoritative. The narrative inherits the machine’s aura of objectivity and computational power while also appearing to possess human-like understanding, reasoning, and communicative competence. This combination is arguably pushing users to build a unique perception of epistemic authority towards AI tools. They are framed and appear both technically superior and cognitively familiar.
If AI is to support rather than disorient human expertise, its authority must be treated as provisional, context-dependent, and empirically demonstrated rather than assumed from general computational performance or human-like presentation. Next, we review the chain from CoT to cognitive deference and suggest that illegitimate grounds for epistemic authority and superiority are what influences over deference, not anthropomorphism.
\section{Overview of CoT to cognitive deference}
In summary, CoT can act as a user-facing explanation trace, that also provides anthropomorphic cues which make the system appear as an intentional agent. Deference becomes likely when this apparent reasoner is also perceived as epistemically authoritative and cognitively superior. Anthropomorphism is a vehicle which makes cognitive reliance through deference/offloading intelligible, while perceived authority and superiority drives reliance. 

One risk is that CoT can make a fallible system or anomalous output appear to satisfy the cues normally reserved for justified trust in expert judgement. The danger arises when this apparent reasoning is taken as evidence of understanding. Moreover, when perspectives of machine superiority amplify the authority of the system’s ‘understanding’. In this case, the reasoning is better described as persuasion rather than potential explanation. 

Cognitive deference becomes problematic when authority and superiority are not granted legitimately. If users mistake CoT for genuine reasoning, they may treat the system as having authority (e.g., expertise, understanding, or judgement) beyond its understood function. Legitimate epistemic authority requires sufficient understanding of the technology’s capacities and limits, and CoT does not meet this standard if it is not a valid or faithful explanation of the process that produced the output (see Section 8). Empirical work on socio-technical bias shows that post-hoc explanations can inflate trust even when incomplete, misleading, or unfaithful to the internal process. CoT is therefore one example of a wider socio-technical bias problem, addressed next.
\section{Cognition deference, the fallacy of composition and anthropomorphic outputs}
Cognitive deference to AI is useful in some contexts and may support more fluid, seamless use of AI in high-stakes decision-making \cite{Hernström2025}. CoT could also be used as a post-hoc tool for giving users greater confidence, perceived certainty, and willingness to defer when the AI appears more rational or reliable than human judgment. Nonetheless, potential harms may result from callous cognition deference. For instance, using an AI tool for medical diagnosis tasks can come at a cost to one’s metacognitive calibration of confidence in their decisions, being overconfident but not accurate or underconfident but accurate \cite{cheatle_accurate_2026}.

An important example of the harms of over-deference is described by the fallacy of composition. People commit a fallacy of composition when they infer that an AI’s strong performance on a narrow task implies broad competence across tasks requiring untested abilities. Over generalising competencies to a model can lead to harmful cognition deference where people rely on some models for inappropriate tasks, such as using a companion chatbot as safe assistance during a mental health crisis \cite{pichowicz_performance_2025}.  This vulnerability of deference is a common tendency in reasoning, and we tend to exhibit such generalisation bias across categories. For instance, success observed in one instance, context, or sub-domain is often taken as evidence of broader competence \cite{hayes_inductive_2018, osherson_category-based_1990}. 

In clinical settings, this may be especially risky because neighbouring tasks can appear superficially similar while relying on different training data or underlying mechanisms with varying constraints. For example, a model that performs well at detecting pneumonia on chest scans may be assumed to possess more general diagnostic competence, even though its success may not transfer to rarer pathologies, atypical presentations, or different imaging populations \cite{ferreira_clinical_2010}. This is a fallacy of composition because competence in one part of the clinical task-space is mistakenly projected onto the whole.

This tendency may be further exacerbated by anthropomorphism, which imports hidden assumptions and misleading associations from human-specific concepts into the interpretation of model capabilities. Expressly, anthropomorphism may act as a moderator for fallacious composition in some cases, as intentional-state inferences may inherently carry-over hidden assumptions that individual task competencies are grounded in more general 'cognitive’ capacities. Many task competencies observed in humans are grounded in cognitive capacities which generalise across a range of task domains. Attributing such cognitive capacities to models could moderate whether one commits the fallacy. Alternatively, if the model is not anthropomorphised but understood in relation to its mechanistic constraints, the generalisation of competencies may be less likely. Its performance is more likely to be treated as task-specific pattern competence rather than as evidence of a broader, general cognitive capacity. 
\section{Potential implications for real-world use: lessons from explainability in clinical contexts}
Clinical contexts provide an important setting for examining post-hoc explanations and cognitive deference because, unlike general uses where evidence remains largely anecdotal, clinical tools are subject to empirical evaluation, oversight, and intervention. Moreover, the increasing use of AI in clinical decision-making raises questions about whether explanations help clinicians critically evaluate recommendations or encourage inappropriate reliance on them.  Explainability (whether through written or visual model outputs) is often proposed as a means of improving transparency, supporting trust and identifying bias. However, Ghassemi et al. argue that commonly used explanation methods may not achieve these objectives for individual decisions and should not substitute for rigorous validation \cite{ghassemi_false_2021}. Though CoT reasoning and post-hoc explanation warrant conceptual distinction \cite{turpin_language_2023}, relevant parallels can be drawn between them as almost all types of model explainability have been empirically demonstrated to influence the weight clinicians assign to AI advice without necessarily improving their ability to distinguish correct from incorrect recommendations \cite{gomez_explainable_2025, jabbour_measuring_2023}. The effects of explanations also depend on users and tasks. Xu and colleagues compared explanation formats, including LLM-generated explanations, in dermatological diagnostic experiments involving lay participants and primary care physicians \cite{xu_divergent_2026}. Presented with these explanations, lay participants exhibited greater automation bias, whereas experienced physicians were more resilient to incorrect AI advice. Gaube et al. similarly examined how clinical expertise and the attributed source of advice shape responses to decision support \cite{gaube_as_2021}. Together, these studies support examining cognitive deference as a conditional response rather than an inevitable consequence of exposure to AI. Here, cognitive deference refers to the clinician assigning greater epistemic weight to an AI recommendation than to their own assessment; whether this is appropriate depends on the quality of the recommendation and the grounds for accepting it. 
\subsection{Can CoT be valid explanation in real world settings?}
The ethical motivation for requiring explanations of specific AI outputs relates to the need for a clinician who is making a decision, to have ‘good reasons’ for that determination. When the reasons rely on technologies, the clinician should sufficiently understand the given technology to allow them to apply their judgement appropriately (the ‘learned intermediary’ rule). London describes how it isn’t the mechanistic understanding of the technology or intervention that matters, because we do not always know precisely how a given intervention (E.g., a drug) produces a specific therapeutic effect \cite{london_artificial_2019}. What we do know is that giving a drug under a given set of circumstances to a specified group of patients will produce a particular therapeutic effect – the legitimate evidence for this belief comes from valid, high-quality clinical studies. London argues that the motivations behind the desirability of explainability misjudges the legitimate empirical foundation for the evidence we need to justify the use of an AI tool \cite{london_artificial_2019}.

Just like with drugs, we may not always have precise understandings of why an AI system produces a specific output. Even the best-performing systems can be brittle and make mistakes. But reliable evidence about their legitimate use can come from well-designed clinical studies that provide evidence that the use of the system causes or is associated with an improvement in patient outcomes or other relevant health system outcomes. 

All users should understand that the AI output can be mistaken, ideally, they know in which kinds of cases that is more likely than in others – a question which is about the measurement of the tool’s prospective performance not a matter of explainability. Clinicians are judged according to the reasonable physician standard \cite{mccradden_explaining_2024}. They are expected to use tools and evidence as a competent physician would under a reasonableness standard, which applies not to traditional automation systems and generative models. When a generative model provides an explanation, it may appear to be reasoning, but this does not mean the system understands or has genuinely justified its answer. If clinicians mistake these explanations for real reasons, they may believe they understand the tool when they do not. This challenges the reasonableness standard because CoT can appear to provide clinical reasoning while actually misleading clinicians away from the patient-specific evidence needed for a reasonable decision.
\section{The future of CoT research}
Importantly, we consider that exploring CoT as explanation is a valid research direction. If we developed the right structural foundation where it was reliable as a means of providing a specific account of the computations behind a particular output or set of outputs, it would be a valid and reliable component of an AI product. However, two issues are left unaddressed, even if this criterion were satisfied: (1) Model explanations can be faithful in explaining an incorrect output – it does not follow that this will ensure the user dismisses the incorrect output. We do not have evidence to suggest that such explainability influences the rejection of incorrect AI outputs; in fact, the opposite is seen, where the presence of specific explanations seems to increase the likelihood that wrong AI recommendations will be followed \cite{jacobs_how_2021}. (2) Even if the explanations are faithful and users were perfectly calibrated in their reliance, the justificatory basis for adequate judgment is still not met, particularly in the context of medicine. This is because judgement is about applying the set of information gathered through all the means and modalities available. For instance, generative outputs cannot form the sole basis for medical judgement alone because modern medical judgment is not grounded in a singular source of information. Frontier AI systems may strengthen the evidentiary component (where we have good clinical evidence to support such an assertion), but they are yet to satisfy the broader requirements of clinical judgment.
\section{Conclusion}
In conclusion, CoT should not be treated as transparent reasoning merely because it takes the linguistic form of reasoning. Its user-facing presentation can supply anthropomorphic cues that make generative systems appear to think, understand, and justify, while wider assumptions about machine objectivity and cognitive superiority make those apparent reasons seem worth deferring to. The risk is not only that users misunderstand how these systems work, but that they foster inappropriate epistemic authority, generalising from apparent linguistic fluency to broader competence and replacing independent judgement with plausible explanation. Attributing human cognitive capacities to systems enables the perception of superior cognition towards AI, leading to a fallacy of composition. In clinical contexts, this is especially concerning because reasonable judgement requires validated evidence, professional expertise, and patient values, not merely a fluent account approximating some steps by which an output may have been produced. Until CoT can be shown to be faithful, useful, and capable of improving rejection of incorrect outputs, it should not be operationalised as explainability. CoT may remain valuable as an internal technical method, but user-facing reasoning should be treated cautiously, actively limited in high-stakes settings, and explicitly addressed in AI literacy as a source of possible persuasion rather than understanding.

\section*{Endmatters}

\subsection*{Generative AI Usage Statement}

The authors did not use generative AI tools during the preparation of this paper. No generative AI tool was used to generate the paper’s ideas, citations, figures or conclusions. The authors take full responsibility for the correctness, originality, integrity, and presentation of all content in the paper.

\subsection*{Competing Interests}

The authors declare no competing interests.

\subsection*{Authorship Contributions}

O.L.: Conceptualisation; Investigation; Project administration; Writing – original draft; Writing – review \& editing.
L.T.: Conceptualisation; Writing- original draft, Investigation; writing- review \& editing
H.T.: Conceptualisation; Writing – original draft; Writing – review \& editing.
C.S.: Conceptualisation; Writing – original draft; Writing – review \& editing.
M.M.: Conceptualisation; Writing- original draft, Investigation; writing- review \& editing

\bibliographystyle{ACM-Reference-Format}
\bibliography{references}

\end{document}